\documentclass[conference]{IEEEtran}

\usepackage{cite}

\usepackage[noend]{algorithmic}
\usepackage{graphicx}
\usepackage{amsmath,amssymb}
\usepackage{algorithm}
\usepackage[skip=10 pt]{caption}
\usepackage[bookmarks=false]{hyperref}
\usepackage{setspace}
\usepackage{booktabs}

\usepackage{verbatim}

\IEEEoverridecommandlockouts

\DeclareMathOperator*{\argmax}{arg\,max}

\usepackage{tikz}
\usepackage{pgfplots}
\usepgfplotslibrary{units}

\IEEEoverridecommandlockouts
\title{Knowledge-Based Biomedical Word Sense Disambiguation with Neural Concept Embeddings \vspace{-6mm}}

\author{AKM Sabbir$^{1}$, Antonio Jimeno-Yepes$^{2}$, and Ramakanth Kavuluru$^{3}$
\thanks{$^{1}$AKM. Sabbir is with the Department of Computer Science,
        University of Kentucky, Lexington, KY, USA.
        {\tt\small akm.sabbir@uky.edu}}%
\thanks{$^{2}$A. Jimeno-Yepes is with IBM Research Australia, Melbourne, VIC, Australia.
        {\tt\small antonio.jimeno@gmail.com}}%
\thanks{$^{3}$R.~Kavuluru is the \textit{\textbf{corresponding author}} and is with the Division of Biomedical Informatics (Department of Internal Medicine) and
the Department of Computer Science, University of Kentucky, Lexington, KY, USA.
        {\tt\small rvkavu2@uky.edu}}%
}

\begin{document}

\maketitle

\begin{abstract}
Biomedical word sense disambiguation (WSD) is an important intermediate task in many natural language processing applications such as named entity recognition, syntactic parsing, and relation extraction. In this paper, we  employ knowledge-based approaches that also exploit recent advances in neural word/concept embeddings to improve over the state-of-the-art in biomedical WSD using the public MSH WSD dataset~\cite{jimeno2011exploiting} as the test set. Our methods involve weak supervision  -- we do not use any hand-labeled examples for WSD to build our prediction models; however, we employ an existing concept mapping program, MetaMap, to obtain our concept vectors. Over the MSH WSD dataset, our linear time (in terms of numbers of senses and words in the test instance) method achieves an accuracy of 92.24\% which is a $3\%$ improvement over the best known results~\cite{yepes2015knowledge} obtained via unsupervised means. A more expensive approach that we developed relies on a nearest neighbor framework and achieves accuracy of 94.34\%, essentially cutting the error rate in half. Employing dense vector representations learned from unlabeled free text has been shown to benefit many language processing tasks recently and our efforts show that biomedical WSD is no exception to this trend.  For a complex and rapidly evolving domain such as biomedicine, building labeled datasets for larger sets of ambiguous terms may be impractical.  Here, we show that weak supervision that leverages recent advances in representation learning can rival supervised approaches in biomedical WSD. However, external knowledge bases (here sense inventories) play a key role in the improvements achieved. 
\end{abstract}


%
\IEEEpeerreviewmaketitle

\section{Introduction}
\label{sec-intro}
Biomedical  natural language processing (NLP) that goes beyond simple text processing is increasingly becoming indispensable to
derive value and insights
from vast amounts of unstructured data generated in the form of scientific
articles~\cite{luo2016bridging,cameron2014graph,kavuluru2012up},
clinical narratives~\cite{ctakes,uzuner20112010,kavuluru2015empirical} and health related
social media posts~\cite{sarker2015utilizing,kavuluru2016toward,paul2015social}. Specialized components including named
entity recognition (NER) programs, syntactic parsers, and relation
extractors form the backbone of many high level information extraction and knowledge discovery applications. For most components
in an NLP application pipeline, there is a clear snowball effect of errors in a component in the beginning of the pipeline leading to
more errors in other subsequent components and the final results of the full application. 

Resolving lexical mentions in text to
correct named entities from a fixed terminology is essential in building an effective NER program, which is typically the next
step after sentence segmentation and part-of-speech tagging in most NLP pipelines. All other downstream components and the
full application suffer if lexical ambiguities are not correctly resolved. Recent research also shows that resolving ambiguities
provides performance gains in information retrieval and search system design~\cite{zhong2012word}. In this effort, we employ
knowledge-based methods, neural concept and word
vectors learned through unsupervised deep learning approaches, and a straightforward nearest neighbor approach to achieve new
state-of-the-art results over a public gold standard dataset~\cite{jimeno2011exploiting} 
in biomedical word sense disambiguation~(WSD).

For this effort, WSD specifically deals with identifying the correct sense of a term, among a set of given candidate senses for that term,
when it is presented in a brief narrative along with its context (surrounding text). For example, consider the ambiguous term `discharge'.
It has two unique senses in biomedicine -- (S1).~The first is the administrative process of releasing a patient from a healthcare facility following an in-patient stay for
some treatment or procedure. (S2).~The second sense pertains to bodily secretions of certain fluids from an orifice or wound.  In our task
the ambiguous term `discharge' is specified along with the sense set \{S1, S2\} and an example context -- ``Low risk patients identified using CADILLAC risk score with STEMI treated
successfully with primary PCI have a low adverse event rate on the third day or later of hospitalization suggesting that an earlier \textbf{discharge} is safe
in properly selected patients.'' Our goal is to identify the correct sense S1 for this specific occurrence of `discharge'. 

Next, we outline the organization of the rest of the paper. In Section~\ref{sec-back}, we discuss earlier efforts in biomedical WSD and recent approaches that incorporate word embeddings. Our main methods that employ concept/word embeddings including the nearest neighbor approach are detailed in Section~\ref{sec-meth}. We then present
our main results and takeaways in Section~\ref{sec-res}. Finally, in Section~\ref{sec-conclude}, we conclude with
some future research directions involving recurrent neural architectures for  biomedical WSD. 

\section{Background and Related Work}
\label{sec-back}
For a thorough overview of approaches to WSD, we direct the readers
to the survey by Navigli~\cite{navigli2009word}, which suggests mainly three categories -- supervised, knowledge-based,
and unsupervised approaches. Supervised approaches for WSD~\cite{zhong2010makes,stevenson2008disambiguation} use a labeled dataset
along with interesting lexical/syntactic features derived from the context around the term to build machine learned models that predict the correct sense
in unseen test contexts.  Knowledge-based approaches~\cite{jimeno2011exploiting,mcinnes2013evaluating} do not use any corpus but solely rely on thesauri
or sense inventories such as
WordNet and the Unified Medical Language System (UMLS) that contain brief definitions of different senses and corresponding synonyms. 
Unsupervised approaches may employ topic modeling~\cite{kim2015link} based methods to disambiguate when the senses are known ahead of time. 
Some unsupervised approaches~\cite{wang2015sense} are often referred to as performing word sense \textit{discrimination} or
\textit{induction} as opposed to disambiguation
because they employ clustering approaches where different clusters are expected to represent the different senses, which are not known a priori.

\subsection{WSD in Biomedicine}
\label{sec-wsdbio}
In biomedicine, knowledge-based word sense disambiguation efforts mostly relied on the UMLS knowledge base~\cite{umlsweb}, which
contains over 3.4 million unique concepts expertly sourced from $\approx$ 200 different terminologies in biomedicine and allied fields. The UMLS is maintained
by the US National Library of Medicine (NLM) and is updated every year to reflect new concepts and other changes. For each concept in the UMLS, there is usually 
a brief definition and sometimes additional relations (both hierarchical and associative)  connecting it with other concepts. Each concept has a
unique ID called the concept unique identifier (CUI), an alphanumeric string that starts with a `C'.
For example, the sense S1 (administrative process) for `discharge'
in Section~\ref{sec-intro} is represented by CUI C0030685 and sense S2 (body substance) is represented by the CUI C0012621.
S1 has a short definition ``The administrative process of discharging the patient, alive or dead, from hospitals or other health facilities''.
For S2 we notice the definition -- ``In medicine, a fluid that comes out of the body. Discharge can be normal or a sign of disease.''
In the MSH WSD dataset that we use in this effort, the candidate senses for each ambitious word are represented in the form of these unique CUIs.
The task is to identify the correct CUI given a particular context (few sentences) containing an ambiguous term. 
For the rest of the manuscript, we use the three terms \textit{CUI}, \textit{concept}, and \textit{sense} synonymously as they refer to the same notion.

Schuemie et al.~\cite{schuemie2005word} present a nice survey of approaches and efforts in biomedical WSD until 2005 including the
well-known NLM WSD dataset~\cite{weeber2001developing}, which has 50 ambiguous terms with 5000 test instances. Disambiguation efforts were also
focused on a small set of 10--15 ambiguous abbreviations~\cite{pakhomov2005abbreviation,xu2012combining} using combinations of supervised and
unsupervised approaches. More recent approaches~\cite{liu2004multi,savova2008word} used supervised models including Naive Bayes, SVMs, logistic
regressors, decision lists with a variety of features using both subsets of the NLM WSD dataset and other smaller datasets.
Berster et al.~\cite{berster2012hyperdimensional} encoded senses, contexts, and ambiguous terms using random indexing
and conducted supervised ten-fold cross validation experiments on the NLM WSD dataset using the binary splatter code method.
McInnes and Pedersen~\cite{mcinnes2013evaluating}
used the network structure of the UMLS (specifically the hypernymic trees) and concept definitions to devise concept relatedness measures which are
in turn used for WSD for the MSH WSD dataset. Chasin et al.~\cite{chasin2014word} demonstrated the
application of topic modeling for a clinical WSD dataset of 50 ambiguous terms curated from the Mayo Clinic~\cite{savova2008word}. Recently, Wang et al.~\cite{wang2016clinical} used an active learning strategy
to involve domain experts in an interactive supervised machine learning framework for biomedical WSD.
Among all the datasets available, the MSH WSD that we use in our current effort is the largest
publicly available dataset~\cite{jimeno2011exploiting} for biomedical WSD (more in Section~\ref{sec-meth})
and also has the least skewed distribution (the average percentage of majority sense is 54\%~\cite{wang2016clinical}).

In a recent approach Jimeno-Yepes and Berlanga~\cite{yepes2015knowledge} used a hybrid approach that combined a knowledge-based component that exploits the UMLS definitions and synonyms
for different concepts
with unlabeled biomedical narratives (from Medline/PubMed)  to derive word-concept probability estimates $P(w|c)$ for any word $w$ and UMLS concept $c$. 
They exploited the Naive Bayes formulation and selected the  correct sense as the CUI $c$ that
maximizes $P(T|c) = \prod_{i} P(w_i|c)$,
where $w_i$ is the $i$-th word in the test context $T$ that contains the ambiguous term. 
With this approach they achieved an accuracy of 89.1\% on the MSH WSD dataset~\cite{jimeno2011exploiting}. 
This result corresponds
to the best performance thus far on the MSH WSD dataset without using supervised models.  Given
we employ this method as a component of our best model, for completeness, we provide its high level summary
in the Appendix. 

In this effort, we use recent advances in neural word embeddings to generate new state-of-the-art results on the MSH WSD dataset achieved
without supervised cross validation experiments.
Our methods can be classified as weakly supervised given we employ the well-known biomedical concept mapping tool MetaMap~\cite{metamap}
to generate concept vectors and employ them in combination with the knowledge-based method from Jimeno-Yepes and Berlanga~\cite{yepes2015knowledge}.

\subsection{Neural Embeddings for WSD}
Neural word representations have been shown to capture both semantic and syntactic information and a few recent approaches
learn word vectors~\cite{bengio2003neural, collobert2008unified, mikolov2013distributed} (as elements of $\mathbb{R}^d$, where $d$ is the dimension)
in an unsupervised fashion from textual corpora. These dense word vectors obviate the sparsity issues inherent to the so called \textit{one-hot} representations
of words\footnote{One-hot representations lead to very large dimensionality (typically the size of the vocabulary) resulting in further issues in similarity computations, a phenomenon
often termed as the \textit{curse of dimensionality} \cite[Chapter 1.4]{bishop2006pattern}}. 
Chen et al.~\cite{chen2014unified} adapted neural word
embeddings to compute different sense embeddings (of the same word) and showed competitive performance
on the SemEval 2007 WSD dataset~\cite{navigli2007semeval}. Disambiguation is achieved by picking the sense that
maximizes the cosine similarity of the corresponding sense vector with the context vector for an ambiguous term. Recently, Iacobacci et al.~\cite{iacobacci2016embeddings}
evaluated and demonstrated the superiority of neural word embeddings as features in supervised WSD models on the same SemEval dataset.

In a very recent effort Pakhomov et al.~\cite{pakhomov2016corpus} use word embeddings (without corpus enhanced concept
embeddings) for the MSH WSD dataset but only report 77\% accuracy although the central aim of their paper is not limited to WSD.
Their approach relies on vectors of words that co-occur with words in the definitions of different senses (CUIs) in the UMLS.  
In our effort, we use a similar framework as Chen et al.~\cite{chen2014unified}  to directly learn sense vectors using
a pure distributional semantics framework that does not rely on word vectors. Additionally, we
employ complementary evidences beyond cosine similarity to achieve further improvements that rival performances typically reported using supervised approaches. 

\section{Our Approach}
\label{sec-meth}
There are 203 ambiguous terms in the MSH WSD dataset~\cite{jimeno2011exploiting} with a total of 424 unique
CUIs (from the UMLS), each of which is a unique sense. Thus, on average, the dataset
has $424/203 = 2.08$ senses. There are a total of 38,495 test instances of contexts (a few sentences) each with one of the 203
ambiguous
terms along with the correct sense (CUI). Besides being the largest biomedical WSD dataset, it also
includes a richer set of ambiguities including 106 ambiguous abbreviations, 88 ambiguous noun phrases, and 9 that
are combinations of both. Due to these features, the NLM encourages researchers to use this
dataset over their older dataset (please see \url{https://wsd.nlm.nih.gov}). 
Our goal is to directly test on this dataset by employing weakly supervised approaches.
For this, we learn vector representations of words and CUIs using well-known approaches that apply deep neural
networks to NLP tasks. 

\subsection{Neural Word and Concept Embeddings}
\label{sec-embed}

We ran the well-known word2vec~\cite{mikolov2013distributed} word embedding
program (the skip-gram model) from Google on over 20 million biomedical citations (titles and abstracts)
from PubMed to obtain word vector representations with a
word window size of ten words and dimensionality $d=300$ with all other parameters set to the default settings.
To learn concept or CUI vectors of the same dimensionality, we curated a dataset of
five million randomly chosen citations (published between 1998 and 2014). For this subset of PubMed, we ran
MetaMap~\cite{metamap} with its WSD option turned on so we obtain unique CUIs for
potential ambiguous terms\footnote{MetaMap uses the UMLS knowledge base of concept 
synonyms along with shallow linguistic parsing to map free text to UMLS CUIs. It also has a WSD option which is based on
concept profiles generated through words co-occurring with different
concepts in biomedical literature~\cite{yepes2012integration}.}. The text was passed through MetaMap two adjacent non-stop words at a time,
to capture as many CUIs as possible. Next, we treated these sequences of CUIs in each citation thus obtained through NER
as a semantic version of the free text corpus. We ran word2vec on this corpus of CUI texts,
just like how we ran it on free text articles with the same parameters. As a result we obtained 300 dimensional dense vectors for
each CUI, including all 424 CUIs corresponding to the 203 ambiguous terms in our test set.
This component of our methodology to derive dense concept vectors involves weak supervision because although
MetaMap with its WSD
option is in and of itself not a powerful solution (see Section~\ref{sec-res}), it nevertheless was useful to learn concept vectors that in turn
helped us achieve state-of-the-art results.
This deep neural network based distributional semantics approach to learning CUI vectors aids in modeling complementary aspects of similarity. This is because we use, as a component, the CUI definition based information via
the word-probability estimate based approach~\cite{yepes2015knowledge} outlined earlier.

\subsection{WSD with Word/Concept Embeddings and Knowledge-Based Approaches}
\label{sec-weak}
Our main idea is that besides comparing pairs of word vectors and concept vectors, we
can also compare a word vector with a concept vector given at a high level there is a direct connection between
words and concepts -- words are often lexical manifestations of high level concepts. 
The fact that we simply replaced word sequences in free text with the corresponding concept sequences
to generate CUI vectors of the same dimensionality as the word vectors also makes it feasible to compare
word vectors and their compositions to concept vectors. As will see in Section~\ref{sec-res}, this intuition
appears to work as well as other state-of-the-art approaches~\cite{yepes2015knowledge}. 

\setlength{\dbltextfloatsep}{8pt plus 1.0pt minus 2.0pt}
\begin{figure*}[ht]
\centering
\includegraphics[scale=0.17]{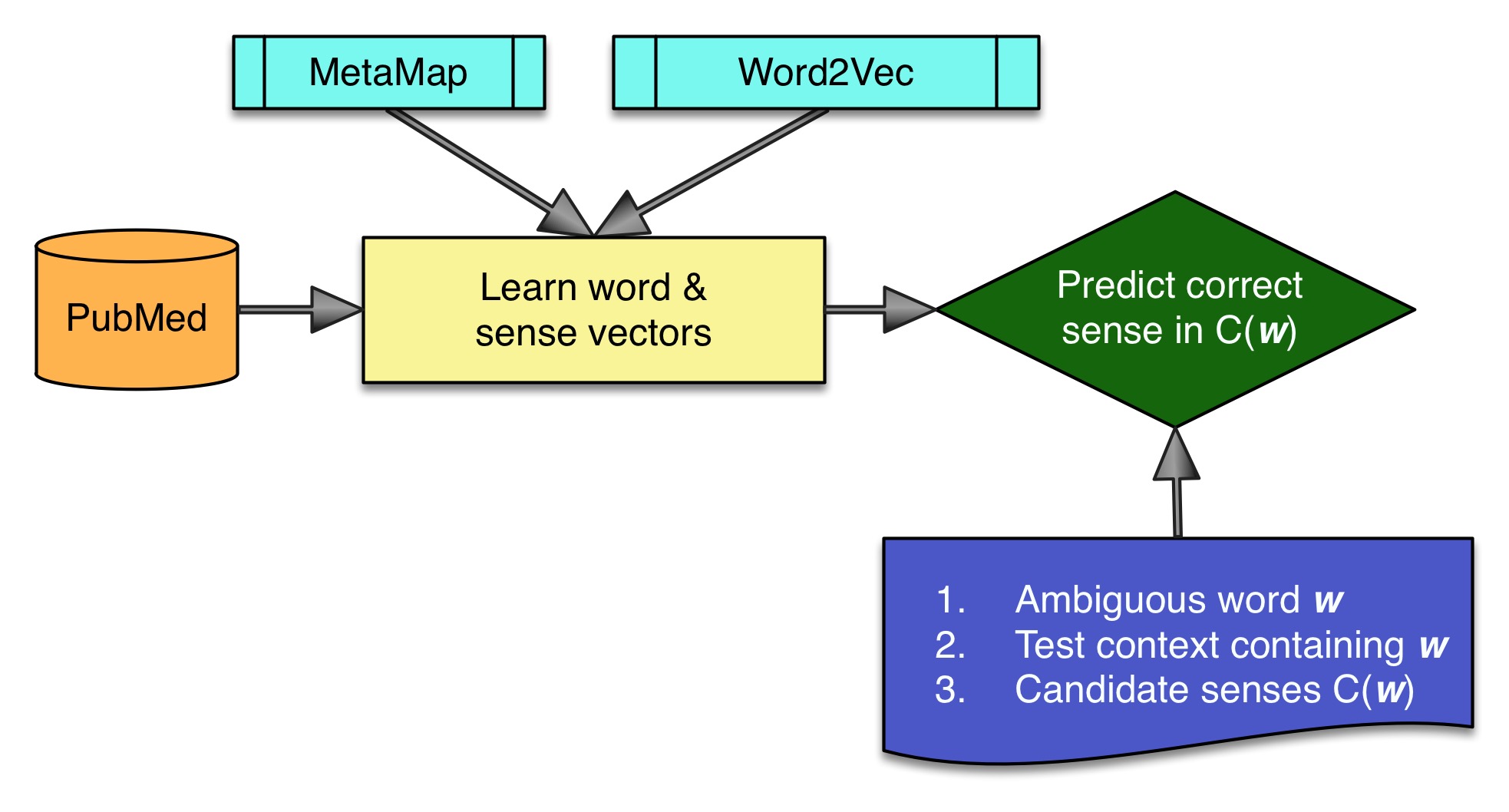}
\caption{Architecture for WSD approaches from Sections~\ref{sec-embed} and ~\ref{sec-weak} \label{fig-arch}}
\end{figure*}

We establish some notation for the rest of the paper. In any WSD problem, a test instance corresponds to a three tuple
$(T, w, C(w))$ where $T$ is a context, typically a few sentences, that contains the ambiguous term $w$ and
$C(w)$ is the set of different senses that $w$ can assume depending upon the context $T$\footnote{In practice,
there might be cases where the context in $T$ is deemed insufficient even for human judges to pick the right sense.
However, for this paper we assess our performance based on MSH WSD dataset where each instance is assigned a unique sense.}. 
Specifically, $C(w)$ in this effort is
the set of different CUIs that capture the different senses for $w$. Our WSD goal is
to construct a function $f(T, w, C(w))$ that maps $T$ to the
CUI $c \in C(w)$ that corresponds to the correct sense in which $w$ was used in $T$.
We have four approaches that apply the embeddings from Section~\ref{sec-embed} to our test set.
We specify them in terms of functions $f^?(T, w, C(w))$ where $?$ indicates symbols that identify the underlying method(s) used, made clear as follows.
\begin{enumerate}[\setlength{\itemsep}{6pt}
\setlength{\labelsep}{6pt}
 \setlength{\IEEElabelindent}{6pt}]
\item Our first approach uses vector cosine similarity with
\[f^c(T, w, C(w)) = \argmax_{c \in C(w)} \cos (\vec{T}_{avg}, \vec{c}), \]
where $\vec{T}_{avg}$ is the average of non-stop words' vectors in the context $T$ 
and $\vec{c}$ is the context vector for $c$. 
This formulation is well-known given cosine similarity is a popular approach to measure semantic similarity of entities (words, concepts, \ldots) represented by the corresponding vectors.

\item Our second approach is based on vector projections with
\[f^p  = \argmax_{c \in C(w)}  \left[ \rho[ \cos (\vec{T}_{avg}, \vec{c})]  \cdot \frac{ \| \mathcal{P} (\vec{T}_{avg}, \vec{c}) \|} {\| \vec{c}\|}  \right], \]
where $\mathcal{P}(\vec{r}, \vec{s})$ refers to the projection of $\vec{r}$ on to $\vec{s}$, $\| \,\, \|$ is the Euclidean norm,
and $\rho$ is the sign function. Using straightforward manipulation based on vector projections in Euclidean spaces~\cite[Chapter 5]{elalg}, we have
\[ \| \mathcal{P} (\vec{T}_{avg}, \vec{c}) \| =  \frac{|\vec{T}_{avg} \bullet \vec{c}|} {\| \vec{c}\|}, \]
which is what we used in our implementation (with $\bullet$ denoting vector dot product). 
 Although $f^c$ (approach one) accounts for the overall
directional similarity (thematic orientation) of the vectors, it does not account for the strength or magnitude of association,
an aspect that seems ignored in others' efforts we reviewed for this paper.  By considering the
vector projection of the context vector onto the CUI vector $\vec{c}$, in $f^{p}$ we also account for the magnitude of
the context vector's projection in relation to that of the CUI vector. The sign function $\rho$ is essentially to account for
situations when $90 < \theta \leq 180$, the angle between $\vec{T}_{avg}$ and $\vec{c}$.

\item Our third approach is based on the first two approaches where we set
\[f^{c,p} = \argmax_{c \in C(w)}  \left[ \cos (\vec{T}_{avg}, \vec{c}) \cdot \frac{ \| \mathcal{P} (\vec{T}_{avg}, \vec{c}) \|} {\| \vec{c}\|}  \right]. \]
We simply incorporate both evidences (magnitude and orientation of association) to compare different CUIs. 

\item Our final approach uses a probabilistic model developed in an earlier effort by Jimeno-Yepes and Berlanga~\cite{yepes2015knowledge} (as outlined in Section~\ref{sec-wsdbio} and elaborated in the Appendix) that selects the $c$ that maximizes $P(T|c)$.  We involve this knowledge-based approach as a third scoring component and set
\begin{align*}
&f^{c,p,k} \\
&= \argmax_{c \in C(w)}  \left[ \cos (\vec{T}_{avg}, \vec{c}) \cdot \frac{ \| \mathcal{P} (\vec{T}_{avg}, \vec{c}) \|} {\| \vec{c}\|} + P(T|c) \right].
\end{align*}
Although there are different ways of combining evidences from multiple sources of predictive information, we rely on this straightforward combination as a form of unsupervised rank aggregation from two different sources. 
\end{enumerate}

The methods discussed thus
far can be summarized using the schematic in Figure~\ref{fig-arch}.

\subsection{WSD with Weak Supervision}
\label{sec-dist}
From methods in Section~\ref{sec-weak}, we have multiple ways of disambiguating CUIs for any ambiguous term given a sample context. 
We exploit them to build a  weakly supervised dataset for the 203 ambiguous terms in our test dataset. 
For each sentence in an independent corpus of biomedical citations
that contains any ambiguous term from our dataset, we employ methods in Section~\ref{sec-weak}
to assign the predicted CUI. Thus we can create a weakly supervised dataset for each ambiguous term
with thousands of examples if we choose a large corpus. These examples can then be used to train traditional discriminative
models or nearest neighbor models. We emphasize here that we are proposing to label arbitrary sentences (not our test sentences)
in an external corpus based on our methods in Section~\ref{sec-weak}. Hence we still have our full MSH WSD dataset to finally
test the approach we propose here with other models in a fair way. 

For the $k$ nearest neighbor ($k$-NN) model, let $\mathcal{D}^w \subseteq \mathcal{D}$ be the set of instances for the ambiguous term
$w$ in the weakly 
supervised dataset $\mathcal{D}$. We rank instances $(D, w, c) \in \mathcal{D}^w $ for a given test instance $T$
based on $\cos (\vec{T}_{avg}, \vec{D}_{avg})$, where $c$ is the sense assigned to $D$ from $C(w)$ based on
methods in Section~\ref{sec-weak}. Let $R_k(\mathcal{D}^w)$ be the set of top $k$ instances in $\mathcal{D}^w$ when
ranked in descending order based on $\cos (\vec{T}_{avg}, \vec{D}_{avg})$.
Now the predicted sense for $T$ is chosen based on 
\[ f^{k-NN} = \argmax_{c \in C(w)}  \left[ \sum_{(D, w, c) \in R_k(\mathcal{D}^w)} \cos (\vec{T}_{avg}, \vec{D}_{avg}) \right]. \]
The expression in the $\argmax$ boils down to summing up the similarities of the test context with those contexts in the training dataset that have the same assigned CUI $c$.
We subsequently pick the particular $c$ that maximizes that summation. Intuitively, our approach
aggregates evidence from training instances that are
semantically most similar to our test instance. The choice of $k$ also plays an important role in the performance of $k$-NN approaches as we
observe in the next section.

\section{Results and Discussion}
\label{sec-res}
\begin{figure*}[t]
\begin{center}
\begin{tikzpicture}[trim axis right, trim axis left, font=\footnotesize]
  \begin{axis}[legend pos=south east, width=1.41\columnwidth,
  height=0.5\textwidth,
    symbolic x coords={20, 50, 100, 200, 300, 500,  1000, 1500, 2000, 2500, 3000, 3500, 4000, 4500, 5000},
    xtick=data,  
    xlabel= Number of nearest neighbors $k$ , xlabel style={yshift=-0.5em},
        ylabel= Accuracy, ylabel style={yshift=-0.5em},
                   yticklabel style={/pgf/number format/fixed, 
                         /pgf/number format/precision=2}]
    \addplot coordinates {
        (20, 90.38)
        (50, 91.7)
        (100,  92.07)
        (200, 92.68)
        (300, 92.87)
         (500, 93.07)
        (1000, 93.56)
        (1500,  93.77)
         (2000,  93.81)
          (2500,  93.98)
            (3000, 94.29)
        (3500,  94.34)
         (4000,  94.17)
          (4500,  93.9)
          (5000,  92.7)
    };
  \end{axis}
\end{tikzpicture}
\end{center}
\caption{Accuracy of the $k$-NN approach with varying $k$ \label{fig-top}}
\end{figure*}
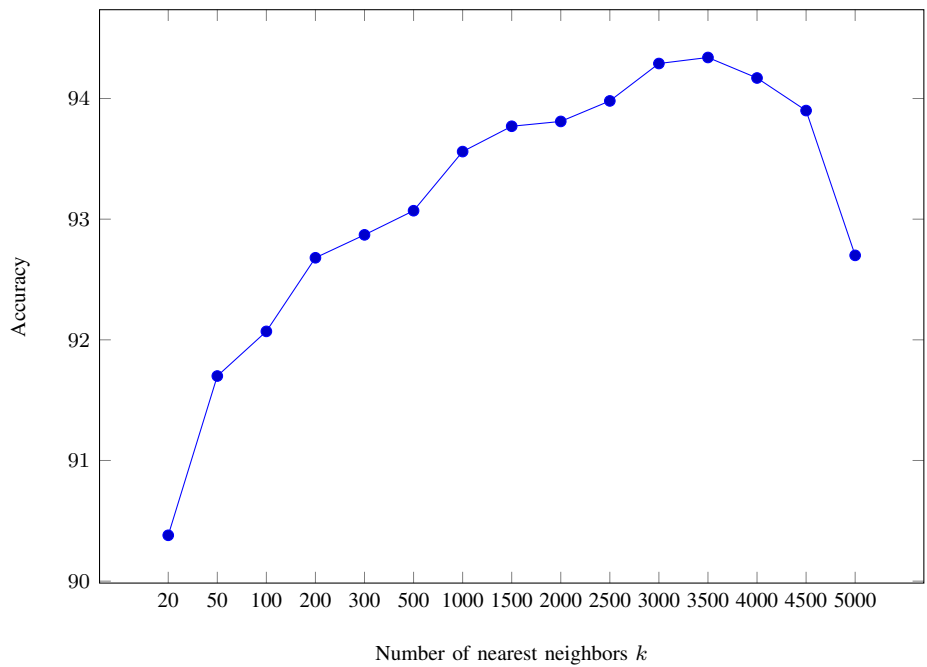

Our results are shown in Table~\ref{tab-res} based on methods introduced in the previous section.
MetaMap does not perform as well
on this dataset (row 1) even with the WSD option achieving an
accuracy\footnote{Accuracy is the ratio of total number of correctly assigned senses to the total of number of occurrences of the 203 ambiguous terms in the MSH WSD dataset.
The usage of accuracy as the evaluation metric is inline with a few prior efforts on biomedical WSD~\cite{mcinnes2013evaluating,pakhomov2005abbreviation,liu2004multi,pakhomov2016corpus} and is
justified~\cite{chasin2014word} given the notions of precision and recall are equivalent to it in this scenario.}
 of 81.77\%. However, it may not be fair to compare MetaMap with our methods given it
does not try to particularly disambiguate our specific 203 terms, for each of which we are already given candidate concepts
that contain the correct sense. 
In row 2 of the table, we show the performance achieved by Jimeno-Yepes and Berlanga
using word-concept probability estimates $P(w|c)$ derived from synonymous names of concepts in the UMLS Metathesaurus. 
\setlength{\intextsep}{6pt plus 1.0pt minus 2.0pt}
\begin{table}[h!]
\renewcommand{\arraystretch}{1.4}
\centering
\caption{Performance on MSH WSD Dataset}
      \begin{tabular}{lc}
        \toprule
        Method & Accuracy \\ \midrule
        MetaMap with WSD & 81.77\% \\
        Jimeno-Yepes and Berlanga~\cite{yepes2015knowledge} & 89.10\% \\ \midrule
        Cosine similarity ($f^c$)  & 85.54\% \\
        Projection length proportion ($f^p$)   & 88.68\% \\
        Combining $f^c$ and $f^p$ ($f^{c,p}$) & 89.26\%  \\ 
        Combining $f^c$, $f^p$, and~\cite{yepes2015knowledge} ($f^{c,p,k}$) & 92.24\%  \\  \midrule
	 Convolutional neural networks & 87.78\%  \\
       $k$-NN with $k=3500$ ($f^{k-NN}$) & 94.34\%  \\ 
        \bottomrule
              \end{tabular}
\label{tab-res}
\end{table}

Rows 3--6 show performances of our methods that leverage neural word/concept embeddings from Section~\ref{sec-weak}.
The cosine similarity and projection approaches both score above 85\% but when used together, they achieve an accuracy of 89.26\%
which is slightly better than the current best result~\cite{yepes2015knowledge} achieved through unsupervised and knowledge-based approaches. Row 6 shows
an accuracy of 92.24\% achieved by our ensemble method that
combines our word/concept vector approach with the knowledge-based  method by Jimeno-Yepes and Berlanga~\cite{yepes2015knowledge}. 
The  time
complexity of these methods is linear in terms of the number of words in the test context $T$ and the number of candidate senses $|C(w)|$,
considering the computation of $\vec{T}_{avg}$ and evaluation of the $\argmax$ expressions for each $c \in C(w)$.

We created a weakly supervised dataset as outlined in Section~\ref{sec-dist} with the same corpus of five million biomedical citations
used for training word and concept vectors (Section~\ref{sec-embed}). From this corpus,
we considered the so called utterances that represent clauses (from the input text) that MetaMap outputs
as distinct fragments with the corresponding CUIs. For each utterance that contains an ambiguous term in our test set,
we apply our best linear method $f^{c,p,k}$ (corresponding to row 6 of Table~\ref{tab-res}) to assign one specific CUI
from all possible candidates. There were seven million such utterances, with an average length of 18 words,
that contained an ambiguous term out of a total of 78 million utterances from the corpus. Given our prior experiences in
convolutional neural networks (CNNs) in biomedical text classification~\cite{rios2015convolutional} that proved superior over
traditional linear classifiers such as support vector machines and logistic regression models, we built 203 multiclass CNN models,
one for each ambiguous term based on this weakly supervised dataset. The configuration of the CNN and its
various hyper parameters were determined as per our prior effort~\cite[Sections 3.2 and 4.2]{rios2015convolutional}.
This setup however resulted in accuracy of 87.78\% which does not match the performance of simpler approaches (rows 4--6  of Table~\ref{tab-res}).

We finally applied the $k$-NN approach outlined in Section~\ref{sec-dist} with the weakly supervised dataset with
the number of nearest neighbors $k \in$ \{20, 50, 100, 200, 300, 500, 1000, 1500, 2000, 2500, 3000, 3500, 4000, 4500, 5000\}. 
The corresponding accuracies are plotted as shown in Figure~\ref{fig-top}. 
We obtained the best accuracy of 94.34\% when $k=3500$ as shown in the last row of Table~\ref{tab-res}. 
Overall, the accuracy rapidly increases as the numbers of neighbors used increase. The gains become smaller as
more neighbors are added, reaching the top score at $k=3500$ after which the accuracy descends abruptly. At $k=5000$, the
accuracy is same as that achieved with $k=300$. This phenomenon is not surprising -- at first more neighbors
contribute to additional evidence, consistency, and robustness against noise in comparing the candidate concepts. However, considering an increasing
number of neighbors at some point also leads to the semantic drift of their content from that of the test context. 
So neighbors ranked further down the list negatively affect the prediction given they are not as related to the test context,
thus lowering overall accuracy. 
We realize that the value of $k=3500$ is specific to this biomedical dataset and that there could be a value 
$3000 < k < 4000$ that achieves a slightly higher accuracy. Our analysis is essentially a proof of concept for the high-level nearly monotonous
nature of performance of $k$-NN based approaches. 
Given that there are over 38,000 test instances in our dataset, we believe $k \approx 3500$ is appropriate in domains with similar characteristics
(e.g., average number of senses per word, distributions of senses, and average length of test contexts). However, researchers
may be able to derive more appropriate $k$ values for their domains if they have access to relevant datasets. 

Finally, it is well
known that $k$-NN approaches are infamous for high test time complexity because of the nearest neighbor search
in high dimensional space.  Our implementation involves cosine similarity computation with all training
instances for the corresponding ambiguous term. In this effort, on average there are nearly 40,000 training instances created through weak 
supervision per ambiguous term.
So given a new test instance $(T, w, C(w))$, cosine similarity (of 300 dimensional vectors)
needs to be computed for the test instance $T$ with about
40,000 contexts to impose the ranking on these potential neighbors. The threshold of a chosen $k$ (say, 3500) can only be
applied after this ranking is created. However, this similarity computation can be parallelized in a straightforward
manner by distributing the similarity computations across multiple processors and pooling the results to incrementally
build the ranked neighbor list. Although real time disambiguation may not be feasible, having the $k$-NN models run overnight
every day to address disambiguation in new articles may be practical. Alternative approaches such as
locality sensitive hashing~\cite{slaney2008locality} that address the dimensionality problems without having to compute
cosine similarities may be helpful to alleviate the situation. Overall, however, it is clear that $k$-NN based
approaches with weakly supervised datasets offer an interesting alternative to purely supervised approaches in biomedical WSD.

\section{Conclusion}
\label{sec-conclude}

Biomedical WSD is an important task with implications for downstream components in NLP applications. 
In this effort, we applied recent approaches in neural word embeddings to construct concept embeddings. Our
linear time method uses these embeddings to combine cosine similarity, projection magnitude proportion, and a prior knowledge-based approach
to produce an accuracy of  92.24\%. This is an absolute 3\% improvement over just using the knowledge-based approach, which generated the
previous best result obtained without supervised learning. Based on predictions from our best linear method, we created a new weakly supervised  dataset
and built a  $k$-NN model that achieves an accuracy of 94.34\%. 

Our results rival performances achieved by
supervised approaches -- the best published supervised result achieves 93\% macro accuracy over ten fold cross validation
experiments on the MSH WSD dataset with the Naive Bayes model~\cite{jimeno2011exploiting}. Based on additional ten fold cross validation 
experiments with support vector machines that use neural word vector features, Jimeno-Yepes~\cite{yepesa2016word} was able to achieve close to 96\% macro accuracy. 
However, we cannot directly compare these cross validation results (from the supervised experiments) to the 94.34\% accuracy we obtained without supervision
in this paper. Specifically, in each iteration, the cross validation framework tests only on one-tenth of MSH WSD dataset by training on the remaining nine folds of the dataset. In our method, we test on the full MSH WSD dataset without using any of it for training. 

Overall, our results in this paper contribute new evidence that dense neural embeddings
function as useful representations of textual data for biomedical NLP applications.
Furthermore, they also showcase the potential of knowledge-based approaches in learning
better dense vector representations (via MetaMap that uses UMLS) and their complementary contributions to WSD tasks.  
We conclude with some limitations and future research directions. 
\begin{itemize}[\setlength{\itemsep}{6pt}
\setlength{\labelsep}{6pt}%
 \setlength{\IEEElabelindent}{6pt}]
\item Although linear method's accuracy is above 92.24\%, there is still room for improvement in terms of incorporating
modifications that account for sense level errors. That is, in addition to accuracy, for each ambiguous term, if we
study the errors (false positives, false negatives) associated with each of its senses, we might be able to modify our approaches to account for any common patterns in which errors manifest. This task involves manual qualitative analysis with over 400 unique senses in the MSH WSD dataset and is a important future research direction. 

\item In Section~\ref{sec-weak}, the test context $\vec{T}_{avg}$ is the vector formed by element-wise
averaging of the word vectors of the corresponding words in the test context $T$. Although averaging is simple and
intuitive, it may not be the best representation of the semantic content of the narrative in the test context.
As such, more powerful
alternatives that can better represent information in the context sentences might be helpful. 
To this end, one option is to directly model paragraphs as fixed size vectors
using a word2vec style unsupervised learning architecture as demonstrated by Le and Mikolov~\cite{le2014distributed}
where paragraph vectors are learned along with word vectors.
\item  A second approach is to consider a weighted average of the word vectors corresponding to
tokens in the context vector where
the weight selected for a word is inversely proportional to its distance from the ambiguous term $w$ in the test context.
Besides word vectors, we can also compute the weighted average of
concept vectors associated with the CUIs (other
than those associated with $w$) in the test context.
The weighted averages of the words and contextual CUIs can then be compared
separately with the candidate concept vectors from $C(w)$ to generate two different scores $\in [0,1]$
whose sum can form the final score to select the correct sense.
\item Both the paragraph vector approach~\cite{le2014distributed} and the weighted averaging approach discussed
earlier do not explicitly model word order when composing test context words to form fixed size vectors that better
capture the semantics of the full context.
Recurrent neural networks (RNNs~\cite[Chapter 3]{DBLP:series/sci/2012-385}), especially with long short-term memory units~\cite{hochreiter1997long}, are a more suitable alternative
for such scenarios but would need training data to set the parameters of the recurrent layer. The dataset created
using weak supervision in Section~\ref{sec-dist} can be used here to estimate RNN parameters corresponding
to the model for each ambiguous term. 
\end{itemize}

\section*{Acknowledgments}
Our work is primarily supported by the National Library of Medicine through grant R21LM012274. 
We are also supported by the
National Center for Advancing Translational Sciences through grant UL1TR001998 and the Kentucky
Lung Cancer Research Program through grant PO2 41514000040001.  The content is solely the responsibility
of the authors and does not necessarily represent the official views of the NIH.
%

\bibliographystyle{IEEEtran}
\bibliography{../../bibs/scoonerdb}

 \newcommand{\noop}[1]{}
\begin{thebibliography}{10}
\providecommand{\url}[1]{#1}
\csname url@samestyle\endcsname
\providecommand{\newblock}{\relax}
\providecommand{\bibinfo}[2]{#2}
\providecommand{\BIBentrySTDinterwordspacing}{\spaceskip=0pt\relax}
\providecommand{\BIBentryALTinterwordstretchfactor}{4}
\providecommand{\BIBentryALTinterwordspacing}{\spaceskip=\fontdimen2\font plus
\BIBentryALTinterwordstretchfactor\fontdimen3\font minus
  \fontdimen4\font\relax}
\providecommand{\BIBforeignlanguage}[2]{{%
\expandafter\ifx\csname l@#1\endcsname\relax
\typeout{** WARNING: IEEEtran.bst: No hyphenation pattern has been}%
\typeout{** loaded for the language `#1'. Using the pattern for}%
\typeout{** the default language instead.}%
\else
\language=\csname l@#1\endcsname
\fi
#2}}
\providecommand{\BIBdecl}{\relax}
\BIBdecl

\bibitem{jimeno2011exploiting}
A.~Jimeno-Yepes, B.~T. McInnes, and A.~R. Aronson, ``Exploiting {MeSH} indexing
  in {MEDLINE} to generate a data set for word sense disambiguation,''
  \emph{BMC bioinformatics}, vol.~12, no. 223, 2011.

\bibitem{yepes2015knowledge}
A.~Jimeno-Yepes and R.~Berlanga, ``Knowledge based word-concept model
  estimation and refinement for biomedical text mining,'' \emph{Journal of
  biomedical informatics}, vol.~53, pp. 300--307, 2015.

\bibitem{luo2016bridging}
Y.~Luo, {\"O}.~Uzuner, and P.~Szolovits, ``Bridging semantics and syntax with
  graph algorithm -- state-of-the-art of extracting biomedical relations,''
  \emph{Briefings in bioinformatics}, p. bbw001, 2016.

\bibitem{cameron2014graph}
D.~Cameron, R.~Kavuluru, T.~C. Rindflesch, A.~P. Sheth, K.~Thirunarayan, and
  O.~Bodenreider, ``Context-driven automatic subgraph creation for
  literature-based discovery,'' \emph{Journal of biomedical informatics},
  vol.~54, pp. 141--157, 2015.

\bibitem{kavuluru2012up}
R.~Kavuluru, C.~Thomas, A.~P. Sheth, V.~Chan, W.~Wang, A.~Smith, A.~Soto, and
  A.~Walters, ``An up-to-date knowledge-based literature search and exploration
  framework for focused bioscience domains,'' in \emph{Proc.~of the 2nd ACM
  SIGHIT Health Informatics Symposium}.\hskip 1em plus 0.5em minus 0.4em\relax
  ACM, 2012, pp. 275--284.

\bibitem{ctakes}
G.~K. Savova, J.~J. Masanz, P.~V. Ogren, J.~Zheng, S.~Sohn, K.~K. Schuler, and
  C.~G. Chute, ``Mayo clinical text analysis and knowledge extraction system
  ({cTAKES}),'' \emph{Journal of the American Medical Informatics Association},
  vol.~17, no.~5, pp. 507--513, 2010.

\bibitem{uzuner20112010}
{\"O}.~Uzuner, B.~R. South, S.~Shen, and S.~L. DuVall, ``2010 i2b2/{VA}
  challenge on concepts, assertions, and relations in clinical text,''
  \emph{Journal of the American Medical Informatics Association}, vol.~18,
  no.~5, pp. 552--556, 2011.

\bibitem{kavuluru2015empirical}
R.~Kavuluru, A.~Rios, and Y.~Lu, ``An empirical evaluation of supervised
  learning approaches in assigning diagnosis codes to electronic medical
  records,'' \emph{Artificial intelligence in medicine}, vol.~65, no.~2, pp.
  155--166, 2015.

\bibitem{sarker2015utilizing}
A.~Sarker, R.~Ginn, A.~Nikfarjam, K.~O'Connor, K.~Smith, S.~Jayaraman,
  T.~Upadhaya, and G.~Gonzalez, ``Utilizing social media data for
  pharmacovigilance: a review,'' \emph{Journal of biomedical informatics},
  vol.~54, pp. 202--212, 2015.

\bibitem{kavuluru2016toward}
R.~Kavuluru and A.~Sabbir, ``Toward automated e-cigarette surveillance:
  Spotting e-cigarette proponents on {T}witter,'' \emph{Journal of biomedical
  informatics}, vol.~61, pp. 19--26, 2016.

\bibitem{paul2015social}
M.~Paul, A.~Sarker, J.~Brownstein, A.~Nikfarjam, M.~Scotch, K.~Smith, and
  G.~Gonzalez, ``Social media mining for public health monitoring and
  surveillance.'' in \emph{Pacific Symposium on Biocomputing.}, vol.~21, 2015,
  pp. 468--479.

\bibitem{zhong2012word}
Z.~Zhong and H.~T. Ng, ``Word sense disambiguation improves information
  retrieval,'' in \emph{Proceedings of the 50th Annual Meeting of the
  Association for Computational Linguistics: Long Papers-Volume 1}.\hskip 1em
  plus 0.5em minus 0.4em\relax Association for Computational Linguistics, 2012,
  pp. 273--282.

\bibitem{navigli2009word}
R.~Navigli, ``Word sense disambiguation: A survey,'' \emph{ACM Computing
  Surveys (CSUR)}, vol.~41, no.~2, p.~10, 2009.

\bibitem{zhong2010makes}
Z.~Zhong and H.~T. Ng, ``It makes sense: A wide-coverage word sense
  disambiguation system for free text,'' in \emph{Proceedings of the ACL 2010
  System Demonstrations}.\hskip 1em plus 0.5em minus 0.4em\relax Association
  for Computational Linguistics, 2010, pp. 78--83.

\bibitem{stevenson2008disambiguation}
M.~Stevenson, Y.~Guo, R.~Gaizauskas, and D.~Martinez, ``Disambiguation of
  biomedical text using diverse sources of information,'' \emph{BMC
  bioinformatics}, vol.~9, no.~11, 2008.

\bibitem{mcinnes2013evaluating}
B.~T. McInnes and T.~Pedersen, ``Evaluating measures of semantic similarity and
  relatedness to disambiguate terms in biomedical text,'' \emph{Journal of
  biomedical informatics}, vol.~46, no.~6, pp. 1116--1124, 2013.

\bibitem{kim2015link}
S.~Kim and J.~Yoon, ``Link-topic model for biomedical abbreviation
  disambiguation,'' \emph{Journal of biomedical informatics}, vol.~53, pp.
  367--380, 2015.

\bibitem{wang2015sense}
J.~Wang, M.~Bansal, K.~Gimpel, B.~D. Ziebart, and T.~Y. Clement, ``A
  sense-topic model for word sense induction with unsupervised data
  enrichment,'' \emph{Transactions of the Association for Computational
  Linguistics}, vol.~3, pp. 59--71, 2015.

\bibitem{umlsweb}
{National Library of Medicine}. (2009) {Unified Medical Language System
  Reference Manual}. \url{http://www.ncbi.nlm.nih.gov/books/NBK9676/}.

\bibitem{schuemie2005word}
M.~J. Schuemie, J.~A. Kors, and B.~Mons, ``Word sense disambiguation in the
  biomedical domain: an overview,'' \emph{Journal of Computational Biology},
  vol.~12, no.~5, pp. 554--565, 2005.

\bibitem{weeber2001developing}
M.~Weeber, J.~G. Mork, and A.~R. Aronson, ``Developing a test collection for
  biomedical word sense disambiguation.'' in \emph{Proceedings of the AMIA
  Symposium}.\hskip 1em plus 0.5em minus 0.4em\relax American Medical
  Informatics Association, 2001, p. 746.

\bibitem{pakhomov2005abbreviation}
S.~Pakhomov, T.~Pedersen, and C.~G. Chute, ``Abbreviation and acronym
  disambiguation in clinical discourse,'' in \emph{AMIA Annual Symposium
  Proceedings}, vol. 2005.\hskip 1em plus 0.5em minus 0.4em\relax American
  Medical Informatics Association, 2005, p. 589.

\bibitem{xu2012combining}
H.~Xu, P.~D. Stetson, and C.~Friedman, ``Combining corpus-derived sense
  profiles with estimated frequency information to disambiguate clinical
  abbreviations,'' in \emph{AMIA Annual Symposium Proceedings}, vol.
  2012.\hskip 1em plus 0.5em minus 0.4em\relax American Medical Informatics
  Association, 2012, p. 1004.

\bibitem{liu2004multi}
H.~Liu, V.~Teller, and C.~Friedman, ``A multi-aspect comparison study of
  supervised word sense disambiguation,'' \emph{Journal of the American Medical
  Informatics Association}, vol.~11, no.~4, pp. 320--331, 2004.

\bibitem{savova2008word}
G.~K. Savova, A.~R. Coden, I.~L. Sominsky, R.~Johnson, P.~V. Ogren, P.~C.
  De~Groen, and C.~G. Chute, ``Word sense disambiguation across two domains:
  biomedical literature and clinical notes,'' \emph{Journal of biomedical
  informatics}, vol.~41, no.~6, pp. 1088--1100, 2008.

\bibitem{berster2012hyperdimensional}
B.-T. Berster, J.~C. Goodwin, and T.~Cohen, ``Hyperdimensional computing
  approach to word sense disambiguation,'' in \emph{AMIA Annual Symposium
  Proceedings}.\hskip 1em plus 0.5em minus 0.4em\relax American Medical
  Informatics Association, 2012, pp. 1129--1138.

\bibitem{chasin2014word}
R.~Chasin, A.~Rumshisky, O.~Uzuner, and P.~Szolovits, ``Word sense
  disambiguation in the clinical domain: a comparison of knowledge-rich and
  knowledge-poor unsupervised methods,'' \emph{Journal of the American Medical
  Informatics Association}, vol.~21, no.~5, pp. 842--849, 2014.

\bibitem{wang2016clinical}
Y.~Wang, K.~Zheng, H.~Xu, and Q.~Mei, ``Clinical word sense disambiguation with
  interactive search and classification,'' in \emph{AMIA Annual Symposium
  Proceedings}.\hskip 1em plus 0.5em minus 0.4em\relax American Medical
  Informatics Association, 2016, pp. 2062--2071.

\bibitem{metamap}
A.~R. Aronson and F.-M. Lang, ``An overview of {MetaMap}: historical
  perspective and recent advances,'' \emph{Journal of the American Medical
  Informatics Association}, vol.~17, no.~3, pp. 229--236, 2010.

\bibitem{bengio2003neural}
Y.~Bengio, R.~Ducharme, P.~Vincent, and C.~Janvin, ``A neural probabilistic
  language model,'' \emph{The Journal of Machine Learning Research}, vol.~3,
  pp. 1137--1155, 2003.

\bibitem{collobert2008unified}
R.~Collobert and J.~Weston, ``A unified architecture for natural language
  processing: Deep neural networks with multitask learning,'' in
  \emph{Proceedings of the 25th international conference on Machine
  learning}.\hskip 1em plus 0.5em minus 0.4em\relax ACM, 2008, pp. 160--167.

\bibitem{mikolov2013distributed}
T.~Mikolov, I.~Sutskever, K.~Chen, G.~S. Corrado, and J.~Dean, ``Distributed
  representations of words and phrases and their compositionality,'' in
  \emph{Advances in Neural Information Processing Systems}, 2013, pp.
  3111--3119.

\bibitem{bishop2006pattern}
C.~M. Bishop, \emph{Pattern recognition and machine learning}.\hskip 1em plus
  0.5em minus 0.4em\relax Springer, 2006.

\bibitem{chen2014unified}
X.~Chen, Z.~Liu, and M.~Sun, ``A unified model for word sense representation
  and disambiguation,'' in \emph{Proceedings of the 2014 Conference on
  Empirical Methods in Natural Language Processing}.\hskip 1em plus 0.5em minus
  0.4em\relax ACL, 2014, pp. 1025--1035.

\bibitem{navigli2007semeval}
R.~Navigli, K.~C. Litkowski, and O.~Hargraves, ``Semeval-2007 task 07:
  Coarse-grained english all-words task,'' in \emph{Proceedings of the 4th
  International Workshop on Semantic Evaluations}.\hskip 1em plus 0.5em minus
  0.4em\relax Association for Computational Linguistics, 2007, pp. 30--35.

\bibitem{iacobacci2016embeddings}
I.~Iacobacci, M.~T. Pilehvar, and R.~Navigli, ``Embeddings for word sense
  disambiguation: An evaluation study,'' in \emph{Proceedings of the 54th
  Annual Meeting of the Association for Computational Linguistics}.\hskip 1em
  plus 0.5em minus 0.4em\relax ACL, 2016, pp. 897--907.

\bibitem{pakhomov2016corpus}
S.~V. Pakhomov, G.~Finley, R.~McEwan, Y.~Wang, and G.~B. Melton, ``Corpus
  domain effects on distributional semantic modeling of medical terms,''
  \emph{Bioinformatics}, p. In Press, 2016.

\bibitem{yepes2012integration}
A.~Jimeno-Yepes and A.~R. Aronson, ``Integration of {UMLS} and {M}edline in
  unsupervised word sense disambiguation,'' in \emph{2012 AAAI fall symposium
  series}, 2012, pp. 26--31.

\bibitem{elalg}
R.~Larson and D.~C. Falvo, \emph{Elementary Linear Algebra}.\hskip 1em plus
  0.5em minus 0.4em\relax Houghton Mifflin Harcourt Publishing Company, 2008.

\bibitem{rios2015convolutional}
A.~Rios and R.~Kavuluru, ``Convolutional neural networks for biomedical text
  classification: application in indexing biomedical articles,'' in
  \emph{Proceedings of the 6th ACM Conference on Bioinformatics, Computational
  Biology and Health Informatics}.\hskip 1em plus 0.5em minus 0.4em\relax ACM,
  2015, pp. 258--267.

\bibitem{slaney2008locality}
M.~Slaney and M.~Casey, ``Locality-sensitive hashing for finding nearest
  neighbors [lecture notes],'' \emph{IEEE Signal Processing Magazine}, vol.~25,
  no.~2, pp. 128--131, 2008.

\bibitem{yepesa2016word}
A.~Jimeno-Yepes, ``Word embeddings and recurrent neural networks based on
  long-short term memory nodes in supervised biomedical word sense
  disambiguation,'' \emph{arXiv preprint arXiv:1604.02506}, 2016.

\bibitem{le2014distributed}
Q.~Le and T.~Mikolov, ``Distributed representations of sentences and
  documents,'' in \emph{Proceedings of the 31st International Conference on
  Machine Learning (ICML-14)}, 2014, pp. 1188--1196.

\bibitem{DBLP:series/sci/2012-385}
A.~Graves, \emph{Supervised Sequence Labelling with Recurrent Neural Networks},
  ser. Studies in Computational Intelligence.\hskip 1em plus 0.5em minus
  0.4em\relax Springer, 2012, vol. 385.

\bibitem{hochreiter1997long}
S.~Hochreiter and J.~Schmidhuber, ``Long short-term memory,'' \emph{Neural
  computation}, vol.~9, no.~8, pp. 1735--1780, 1997.

\end{thebibliography}

\section*{Appendix}
As a component of main methods in Section~\ref{sec-weak}, we use the knowledge-based approach developed by Jimeno-Yepes and Berlanga~\cite{yepes2015knowledge} which
we briefly discussed in Section~\ref{sec-wsdbio}. Here we give some additional details. The main idea is to model $P(c|T)$, probability of CUI $c$
given a context $T$. If this is estimated accurately, our WSD solution is to simply pick the candidate sense $c$ that maximizes $P(c|T)$. Using Bayes theorem, we have
\begin{equation*}
P(c|T) = \frac{P(T|c) P(c)}{P(T)} \propto P(T|c) = \prod_{w_{j} \in T} P(w_{j}|c),
\end{equation*}
with the naive assumption of independence of tokens $w_j$ that constitute the context $T$ given the sense $c$. 
So our solution now depends on estimating the word-concept probabilities $P(w|c)$ for any word $w$ and CUI $c$. The rest of this appendix outlines how 
Jimeno-Yepes and Berlanga accomplish that.
   
A straightforward first cut to obtain $P(w|c)$ is to simply model it as the maximum likelihood estimate
\begin{equation*}
P(w|c) = \frac{count(w, c)}{\sum_{w' \in \, lex(c) }{count(w',c)}},
\end{equation*}
where $lex(c)$ is the synonymous name set of $c$ in the UMLS. Instead of limiting the search
of $w$ to the lexical space of $c$, they propose to extend it to lexical spaces of concepts that are 
related to $c$ based on the UMLS relations available as part of the MRREL file in the Metathesaurus~\cite{umlsweb}. That is, we
now have $P_k(w|c_0)$, which denotes the probability of $w$ being selected for the set of concepts
$R_k(c_0)$ that are $k$ hops away from the original concept $c_0=c$. Specifically, they estimate
\begin{align*}
&P_{k}(w|c_{0}) = \sum_{c_{k} \in  R_{k}(c_{0})}{P_{k}(w,c_{k},c_{k-1},\ldots|c_{0})}\\
&=\frac{\sum_{c_{k} \in R_{k}(c_{0})}{P_{k}(w,c_{k},c_{k-1},\ldots,c_{0})}}{P(c_{0})}\\
&= \frac{\sum_{c_{k} \in R_{k}{(c_{0})}}P_{0}{(w|c_{k})} \prod_{l=0,\ldots, k-1}{P(c_{l+1}|c_{l}) P(c_0)}}{P(c_{0})}\\
&= \sum_{c_{k} \in R_{k}(c_{0})}{ P_{0}(w|c_{k})} \prod_{l=0, \ldots ,k-1}{P(c_{l+1}|c_{l})},
\end{align*}
where Markov assumption is used for estimating $P_k(w, c_k, \ldots, c_0)$ in terms of
traversal probabilities, $P(c_{l+1}|c_{l})$, of hopping from concept $c_l$ to $c_{l+1}$ in the UMLS relation graph. This is mathematically estimated as
\begin{equation*}
P(c_{l+1}|c_{l}) = \frac{|r(c_{l+1},c_{l})|}{|r(*,c_{l})|},
\end{equation*}
with $r(c_1, c_2)$ denoting the number of UMLS relations connecting $c_1$ and $c_2$ and the denominator indicating the number of relations where
$c_l$ participates. 

The word concept probabilities $P_j(w|c)$ obtained at different values of $j=0, \ldots, l$ are finally combined using 
a linear combination to estimate
\begin{eqnarray*}
P(w|c) = \sum_{j=0, \ldots,l}{\alpha_{j}P_{j}{(w|c)}} \quad \mbox{where} \quad
\alpha_{0}, \ldots, \alpha_{l}  > 0 
\end{eqnarray*}
\[  \mbox{and}  \quad \sum_{j = 0, \ldots, l}{\alpha_{j} = 1}. \]
They start with each $\alpha_j =1/l$ with $l$ being the number of hops considered and update them using
expectation-maximization, details of which are presented in their paper~\cite[Section 3.3]{yepes2015knowledge}.

\end{document}